\documentclass{amsproc}
\usepackage[utf8]{inputenc}
\usepackage[margin=1in]{geometry}

\usepackage{tabularx}
\usepackage{longtable}
\usepackage{booktabs}
\usepackage{subfigure}
\usepackage{graphicx}

\usepackage{xcolor}
\definecolor{darkgreen}{rgb}{0.0, 0.5, 0.0}

\usepackage{hyperref}

\begin{document}


\title{Language models and Automated Essay Scoring}

\author{Pedro Uria Rodriguez, Amir Jafari and Christopher M. Ormerod}

\subjclass[2010]{Primary }


\begin{abstract}
In this paper, we present a new comparative study on automatic essay scoring (AES). The current state-of-the-art natural language processing (NLP) neural network architectures are used in this work to achieve above human-level accuracy on the publicly available Kaggle AES dataset. We compare two powerful language models, BERT and XLNet, and describe all the layers and network architectures in these models. We elucidate the network architectures of BERT and XLNet using clear notation and diagrams and explain the advantages of transformer architectures over traditional recurrent neural network architectures. Linear algebra notation is used to clarify the functions of transformers and attention mechanisms. We compare the results with more traditional methods, such as bag of words (BOW) and long short term memory (LSTM) networks.
\end{abstract}
\maketitle

\section{Introduction}

Automated essay scoring (AES) is the use of some statistical model to assign grades to essays in an educational setting. These engines were initially used to reduce the cost of essay scoring \cite{Page1967, Page1968}. Aside from cost effectiveness, AES is considered to be inherently more consistent and less biased than human raters. We can compare the performance of an AES engine with the performance of human raters using inter-rater reliability (IRR) statistics \cite{IRR}. Recently, an AES engine with above human performance was presented in \cite{alikaniotis2016automatic} based on an engine in which experts carefully engineered a set of features. AES has been a subject of a number of recent works by Sakaguchi \cite{sakaguchi2015effective}, Shermis and Hammer \cite{Shermis2015}, and Yannakoudakis \cite{yannakoudakis2015evaluating}.

The function of AES is essentially one of classification, where neural networks are associated with almost all the current state-of-the-art results. Feedforward (static) neural networks are a class of powerful nonlinear statistical models capable of modelling complex relationships between the input space and a set targets \cite{hagan1994training}. Many of these Feedforward neural networks are known as Convolutional Neural Networks (CNN)'s which are ubiquitously used in image classification tasks \cite{imagenet}. These nonlinear models are fit to a set of training data using backpropagation and a variety of optimization algorithms. Recently very efficient deep neural net model architectures have been used to compute the vector representation of words and/or subwords called embeddings \cite{mikolov2013efficient}. These models are used heavily in Natural Language Processing (NLP) tasks to convert words and/or subwords to vectors in a meaningful manner that has been shown to preserve semantic information.

We also consider AES to be an area of NLP in which another type of dynamic network is ubiquitously used. These dynamic networks are mostly called Recurrent Neural Networks (RNN)'s and are powerful tools used to model and classify data that is sequential in nature. These types of networks have been used in engineering and science in the identification and modeling of complex systems \cite{JAFARI2018312}. Using an embedding we may convert a sequence of words into a sequence of vectors that has preserved the semantic information. RNN's, in combination with embeddings, have many applications in NLP tasks like sentiment analysis, topic labeling, language detection and machine translation \cite{bahdanau2014neural}. In recent years, researchers have applied RNNs and Deep Neural Nets to AES.

In cases where there are a very large number of student essays, grading can be a very expensive and time consuming process. Since scoring essays is a part of the student assessment process that is conducted by almost all educational testing agencies, there are many AES engines being used in large-scale formative and summative assessment \cite{Shermis2015}. The core idea of essay scoring is to evaluate an essay with respect to a rubric which may depend on traits such as the use of grammar, the organization of the essay in addition to topic specific information. An AES engine seeks to extract measurable features which may be used to approximate these traits, hence, deduce a probable score based on statistical inference. A comprehensive review of AES engines in production featured in the work of Shermis et al. \cite{Shermis2015}.

In 2012, Kaggle released a Hewlett Foundation sponsored competition under the name  ``Automated Student Assessment Prize'' (ASAP). Competitors designed and developed statistical AES engines based on techniques like Bag of Words (BOW) in combination with standard Machine Learning (ML) algorithms to extract important features of student responses that correlated well with scores. Subsequent works applied RNN-based engines in combination with word embeddings to the Kaggle AES dataset \cite{lstm}. This dataset and these results provide us with a benchmark for AES engines and a way of comparing current state-of-the-art neural network architectures against previous results.

Since there exists an abundance of unlabeled text data available, researchers have started training very deep language models, which are networks designed to predict some part of the text (usually words) based on the other parts. These networks eventually learn contextual information. By adapting these language models to predict labels instead of words or sentences, state-of-the-art results have been achieved in many NLP tasks. Many of these models (see \cite{opengpt2, bert, xlnet, elmo}) are built from layers of Transformers which utilize attention to find the most relevant features to perform a particular task \cite{elmo}. We concentrate on two such models; the Bidirectional Encoder Representations from Transformers (BERT), introduced in \cite{bert}, and XLNet, which is a variation of the BERT model \cite{xlnet}.

\section{Automated Essay Scoring}

In this section, we discuss the task of producing an AES engine. This includes the data collection, how we train the models and how we evaluate an AES engine. We include a brief description of some of the standard IRR statistics in the literature used in the context of evaluating models \cite{IRR, williamson2012framework}.

The first step in producing an AES engine is data collection. Typically, a large sample of essays is collected for the task and scored by expert raters. The raters are trained using a holistic rubric specifying the criteria each essay is required to satisfy to be awarded each score. Exemplar essays are used to demonstrate how the criteria is to be applied. A holistic rubric may take into account a number of factors such as grammar, spelling, organization, clarity and cohesion \cite{rubrics}. Since these essays are the result of specific prompts shown to students, the rubric may include prompt specific information. The training material for the Kaggle AES dataset was made publicly available. To evaluate the efficacy of an AES engine, we require that every essay is scored by (at least) two different raters. Other quality control mechanisms like resolution reads and targeted backreads help improve the quality of the data \cite{williamson2012framework}.

Once the collection of essays is scored, we divide the essays into three different sets; a training set, a test set and a validation set. From a classification standpoint, the input space is the set of raw text essays while the targets for this problem are the human assigned labels. The goal of an AES engine use and evaluate a set of features of the training set, either implicitly or explicitly, in a manner that the labels of the test set may be deduced as accurately as possible using statistical inference. Ultimately, if the features are appropriate and the statistical inference is valid, the AES engine assigns grades to essays statistically similarly to how a human would on the test set. Once the hyperparameters are optimized for the test set, the engine is applied on the validation set.

In the case of the ASAP data, two raters were used to evaluate the essays. We call the scores of one reader the initial scores and the scores of the second reader the reliability scores. There are two main metrics used to evaluate the agreement between two sets of scores; the exact agreement (accuracy), which measures when two scores agree, and the quadratic weighted kappa (QWK) statistic \cite{cohen1960coefficient} or Cohen's Kappa Score. The QWK of two sets of scores is defined as follows:

\begin{align}
\textbf{K} &= 1 - \dfrac{\sum_{i,j} w_{i,j} x_{i,j} }{\sum_{i,j} m_{i,j} x_{i,j}}\label{eq_m1}\\
m_{i,j} &= x_{i,j}(1-x_{i,j})\\
w_{i,j} & = 1- \dfrac{(i-j)^2}{(k-1)^2}
\end{align}
where $k$ is the number of classes and $x_{i,j}$ is the probability of score $i$ receiving score $j$. The original Cohen's Kappa Score is defined as
\begin{equation}\label{eq_0}
\textbf{K} = \cfrac{p_{o}- p_{e} }{1-p_{e}}
\end{equation}
\noindent where $p_{o}$ is the relative observed exact agreement among raters (i.e., accuracy), and $p_{e}$ is the hypothetical probability of chance agreement, using the observed data to calculate the probabilities of each observer randomly seeing each category \cite{cohen1960coefficient}. The QWK has the property that $\textbf{K}=1$ if the raters are in complete agreement. The QWK captures the level of agreement above and beyond what would be obtained by chance and weighted by the extent of disagreement. Furthermore, in contrast to the accuracy, QWK is statistically a better measurement for detecting disagreements between raters since it depends on the entire confusion matrix, not just the diagonal entries. Typically, the QWK between two raters is also used to measure the quality or subjectivity of the data used in training.

We may evaluate an AES engine on the ASAP dataset and compare the engine with a human rater by training an engine on the initial scores and showing that the scores predicted by the engine are in greater agreement with the intial scores than the reliability scores on the validation set. We used the same 5-fold cross validation splits found in \cite{lstm} where each of the five splits used 60 percent of the data as training data, 20 percent as a test set and 20 percent as a validation set. We also considered hyperparameter tuning at a level in which the very structure of the network was altered.

Automated Essay Scoring is one of the more challenging tasks in NLP. The challenges that are somewhat distinct to essay scoring relate to the length of essays, the quality of the language/spelling and typical training sample sizes. Essays can be long relative to the texts found in sentiment analysis, short answer scoring, language detection and machine translation. Furthermore, while many tasks in NLP can be done sentence by sentence, the length and structure of essays often introduces longer time dependencies which requires more data than typically available. The amount of data is often restricted due to the expense of hand-scoring. The longer the essay, the more difficult for Neural Network models to keep the information from beginning of the essay in the network. This results in convergence issues or low performance. These are in addition to typical challenges of NLP such as the choice of embedding, different contextual meanings of words and the choice of ML algorithms.

A variety of models have been introduced over the last 50 years in essay scoring \cite{PEG}. These models started with statistical models using the Bag of Words (BOW) method with logistic regression or other classifiers, SVD methods for feature selection and probabilistic models like Naive Bayes or Gaussian models. In using Neural Network models, we are required to choose an appropriate embedding \cite{embedding}. An embedding may be between characters, words, subwords or sentences into some real $n$-dimensional space that somehow preserves the usage/semantics \cite{embedding}. This converts a text into a sequence of vectors in $n$-dimensional space which may be modeled using RNNs like LSTM Networks\cite{hochreiter1997long} and GRU Networks \cite{bahdanau2014neural} or with CNNs. The gating mechanisms, such as those found in GRUs and LSTM units, mitigate the issue of long term dependencies to some degree, however, it has been shown that long term dependencies are more effectively accommodated for by attention mechanisms \cite{bert}. Recently people have started to combine these algorithms with each other in order to improve the results.

At a word level, if a word is misrepresented or misspelled the embedding of that token results in an inconsistent input that is being used to train the NN models leading to poor extrapolation. Standard algorithms for correcting words may suggest words that do not fit into the context. The language models in question are masked word models \cite{opengpt2, bert, xlnet, elmo} which seeks to guess a selection of missing words better than standard algorithms by incorporating context in three different ways. These models use three different embeddings; a word/subword embedding, a sentence embedding and a positional embedding that encodes the position of each word. The probably masked words are calculated by using context at a word and sentence level. By modelling sentences, these models possess much more information than typically available using typical word embeddings.

Neural networks are inherently non-linear and continuous models, however, to approximate a discrete scoring rubric, a series of boundaries is introduced in the output space that distinguish the various scores. When the output lies close to the boundaries between scores it is difficult for the models to pick a score correctly. Ideas of committee (or ensemble) of networks by taking a majority vote or the mean will be discussed in later sections.

\section{Models}

In this section we will go into some detail regarding some of the major methods used develop AES engines \cite{Shermis2015}. We start with the BOW method in which the features are explicitly defined. We then go on to describe RNN approaches. In particular, we will review how the gating mechanism in layers of LSTM units allow for long term dependencies. The Multi Layer Perceptron and its variations are classified as static network and networks that have delays are also considered RNNs. Lastly, we elucidate the structure and function of the language models featured in this paper.

\subsection{BOW}

For ML algorithms, we mostly prefer to have well defined fixed input and targets. An issue with modeling text data is that it is usually very messy and some techniques are required to pre-process it into useful inputs and targets to feed to ML algorithms. Texts needs to be converted to numbers that we can use in machine learning as proper input and labels. Converting textual data to vectors is called feature extraction or feature encoding. A bag of words (BOW) model is a technique to extract features from text and use them for modeling. The method is very simple:

\begin{enumerate}
 \item Find all occurrence of words within a document.
 \item Find a unique vocabulary of words.
 \item Then form the vector that represents the frequency of each word.
 \item Each dimension of the vector represents the number of counts (occurrence).
 \item Remove dimensions associated with very high frequency words.
 \item We use term frequency (TF) (take the raw frequency and divide to max frequency).
 \item We use inverse document frequency (IDF) (log of documents counts to the length of all the documents has has the term)
 \item By multiplying the TF and IDF, we get (TF-IDF) to reduce the most important words.
 \item Normalize the TF-IDF vectors.
\end{enumerate}

The BOW model is completed and each essay is associated with a single vector and the set of vectors with a particular label may be classified by some traditional classifier. We should note that the BOW model will not consider the order of the words and that in each bag it finds the words that have the most textual information.

\subsection{LSTM}
The output of an RNN is a sequence that depends on the current input to the network but also on the previous inputs and outputs. In other words, the input and output can be delayed and we can also use the state of the network as input. Since these networks have delays, they operate on a sequence of inputs in which the order is important.

\begin{figure}
\centering
\subfigure[Basic unit]          {\label{f1-a}   \includegraphics[width=2in, height=2in]{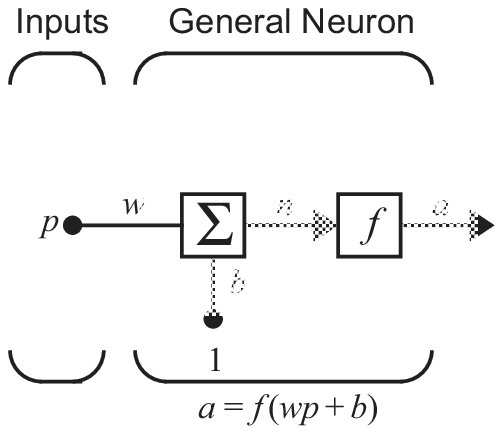}}
\subfigure[MLP]                 {\label{f1-b}   \includegraphics[width=3.5in, height=2in]{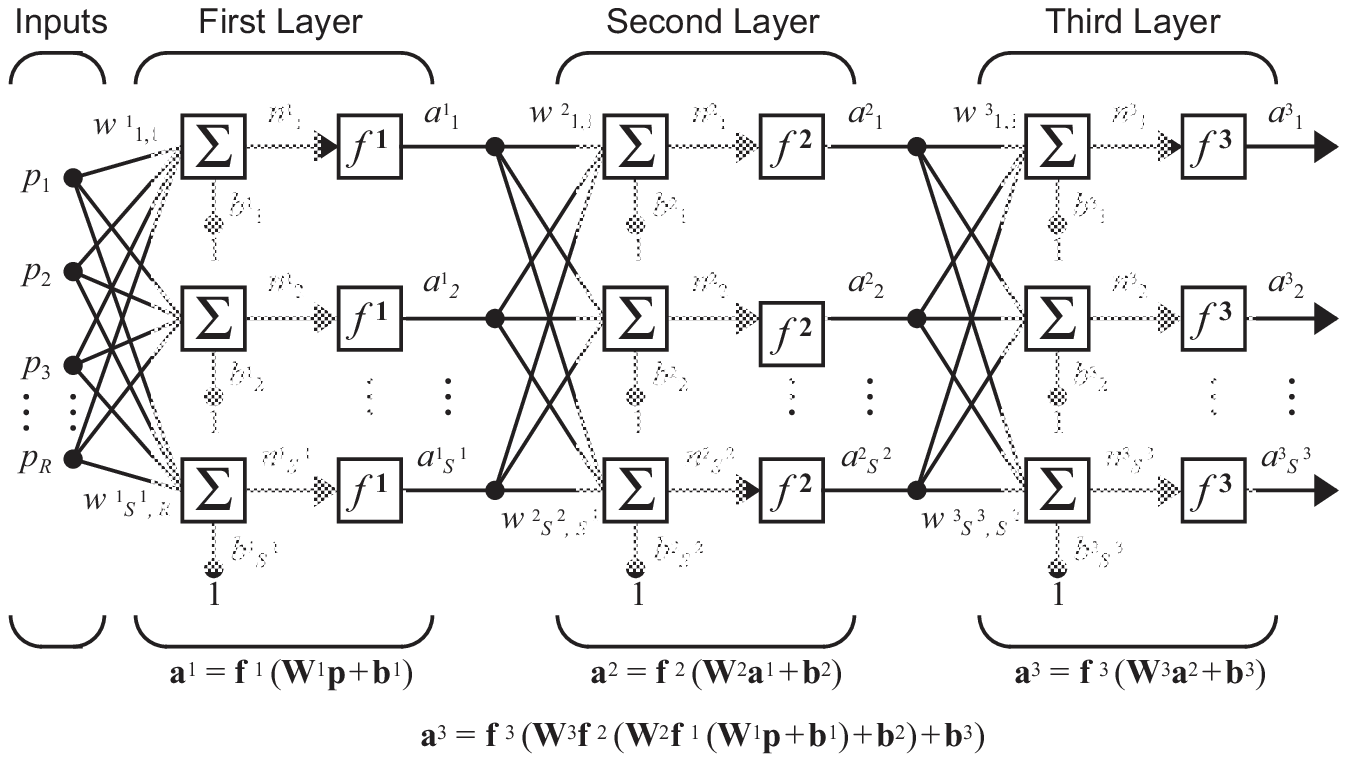}}
\label{f1}
\caption{Artificial Neural Networks Structure \cite{NND}}
\end{figure}

An RNN can be a purely Feed Forward network with delays in the inputs or they can have feedback connections with the output of the network and/or the state of the network. A variety of recurrent units, which are used to build RNNs, are available like LSTM \cite{hochreiter1997long}, GRU\cite{chung2015gated}, ELMAN, NARX \cite{jafari2015enhanced} and Focused Delay Networks. In this section we are going to discuss networks of LSTM units. In order to do so, let us introduce a general notation by describing the most basic unit that makes up all ANNs, that is, the (artificial) neuron, shown in Figure \ref{f1-a}.

A scalar \textit{input} $p$ is multiplied by a parameter $w$, called \textit{weight}, and the result is added to another parameter $b$ called \textit{bias}. Their sum ($n$, the \textit{net input}) goes into a (usually) nonlinear activation function $f(x)$ to get the neuron output $a$. By updating the values of $w$ and $b$ through an iterative optimization algorithm called \textit{Gradient Descent}, this single neuron can find the best parameters that fit the neuron equation (with a set transfer function) to any two-dimensional data. In other words, this single modular unit can map input data to the target and approximate the underlying function. By assigning a different weight to each input dimension, a single neuron can be extended to model $N$-dimensional data. In this case, both $\mathbf{p}$ and $\mathbf{w}$ are $N$-dimensional vectors, and the neuron output equation is

\begin{equation}\label{eq_1}
a = f(\mathbf{W} \cdot \mathbf{p} + b).
\end{equation}

By combining multiple neurons together, and stacking multiple layers of these neurons, a Multi-Layer Perceptron (MLP) is formed \ref{f1-b}. The super script number shows the layer numbers. For example, the forward calculation of the three layers shown in the figure is

\begin{gather}\label{eq_2}
\textbf{a}^{1} = \textbf{f}^{1}(\textbf{W}^{1} \textbf{p} + \textbf{b}^{1}),\\ \nonumber
\textbf{a}^{2} = \textbf{f}^{2}(\textbf{W}^{2} \textbf{a}^{1} + \textbf{b}^{2}),\\ \nonumber
\textbf{a}^{3} = \textbf{f}^{3}(\textbf{W}^{3} \textbf{a}^{2} + \textbf{b}^{3}).
\end{gather}

We want to introduce the neural network framework that we will use to represent general recurrent networks. We added new notation that we have used to represent MLP, therefore we can conveniently represent networks with feedback connections and tapped delay lines.

The net input $n^{m}(k)$ for layer $m$ of an RNN can be computed as follows:

\begin{equation}
\textbf{n}^{m}(k)=\sum_{l\in L^{f}_{m}}\sum_{d\in DL_{m,l}}\textbf{LW}^{m,l}(d)\textbf{a}^{l}(k-d)
\nonumber
\end{equation}
\begin{equation}
+\sum_{l\in I_{m}}\sum_{d\in DI_{m,l}}\textbf{IW}^{m,l}(d)\textbf{p}^{l}(k-d)+\textbf{b}^{m},
\label{eq:net_int}
\end{equation}
where $\textbf{p}^{l}(k)$ is the $l$-th input to the network at time $k$, $\textbf{IW}^{m,l}$ is the input weight between input $l$ and layer $m$, $\textbf{LW}^{m,l}$ is the layer weight between layer $l$ and layer $m$, $\textbf{b}^{m}$ is the bias vector for layer $m$, $DL_{m,l}$ is the set of all delays in the tapped delay line between layer $l$ and layer $m$, $I_{m}$ is the set of indices of input vectors that connect to layer $m$, and $L^{f}_{m}$ is the set of indices of layers that connect directly forward to layer $m$. The output of layer $m$ is
\begin{equation}
\textbf{a}^{m}(k)=\textbf{f}^{m}(\textbf{n}^{m}(k)),
\label{eq:net_out}
\end{equation}
for $m=1,\; 2,\; \cdots,\; M$, where $\textbf{f}^{m}$ is the transfer function at layer $m$. The set of $M$ paired equations (\ref{eq:net_int}) and (\ref{eq:net_out}) describes the general RNN. RNN can have any number of layers, any number of neurons in any layer, and arbitrary connections between layers (as long as there are no zero-delay loops) \cite{NND}.

\begin{figure}
    \centering
    \centerline{\includegraphics[width=6in, height=3.5in]{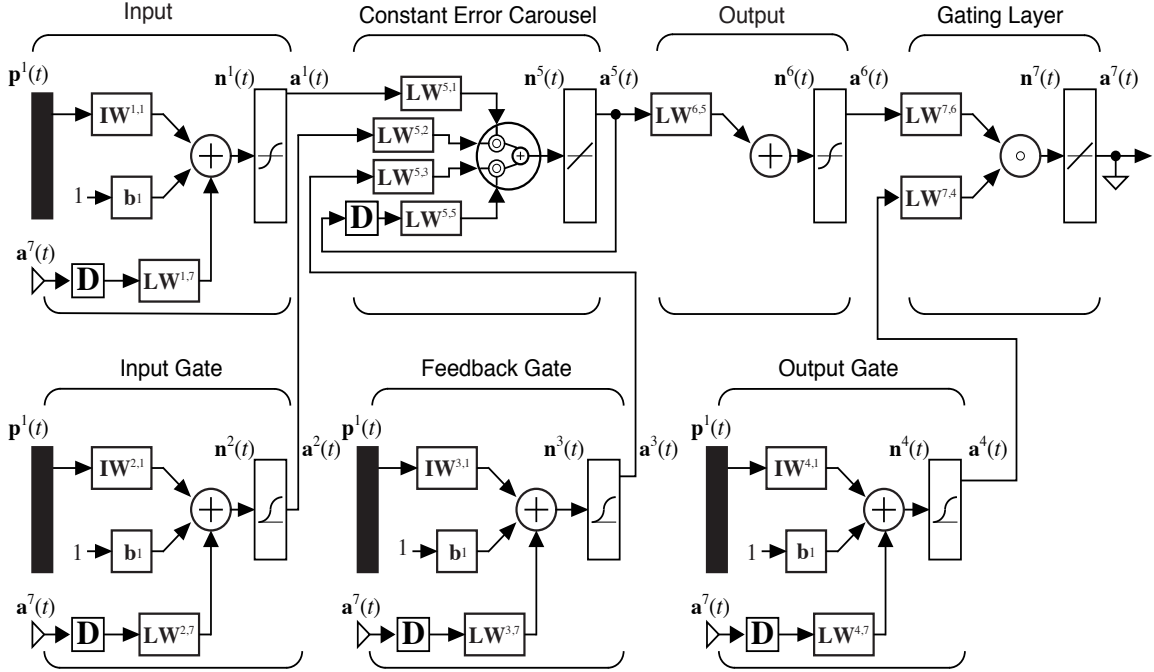}}
    \caption{Long Short Term Memory}
    \label{LSTM}
\end{figure}

Training RNN networks can be very complex and difficult. The key issues that may arise are Vanishing Gradients \cite{hochreiter1997long}, Exploding Gradient and instability \cite{pascanu2013difficulty}. Many architectures are proposed to deal with these issues. Long Short Term Memory (LSTM) network is one of these network architectures \cite{hochreiter1997long} that has recently become very popular. They key concept in LSTM is we would like to predict responses that may be significantly delayed from the corresponding stimulus. For example, words in a previous paragraph can provide context for a translation, therefore the network must enable this possibility to have long term memory.

Long term memories are the network weights and short term memories are the layer outputs. We need a network which has long and short term memory combined. In RNNs, as the weights change during training, the length of the short term memory will change. It will be very difficult to increase the length if the initial weight does not produce a long short term memory. Unfortunately, if the initial weight produces a long short term memory, the network can easily have unstable outputs. To maintain a long term memory, we need to have a layer called Constant Error Carousel (CEC). This layer has a feed back matrix $\textbf{LW}^{1,1}$ to have some eigenvalues very close to one shown in Figure \ref{LSTM}. This has to be maintained during training or the gradients will vanish. In addition to ensure long memories, the derivative of the transfer function should be constant. Therefore, we need to set $\textbf{LW}^{1,1} = \textbf{I}$ and use a linear transfer function.

Now, we do not want to indiscriminately remember everything. Thus, we need to create a system that selectively picks what information to remember. The solution, outlined in \cite{hochreiter1997long} is a gating mechanism in which gates act like switches that operates on input, the CEC layer and the output layer. The input gate will allow selective inputs into CEC, a feedback or forget gate will clear CEC, and the output gate will allow selective outputs from CEC. Each gate will be a layer with inputs from gated outputs and the network inputs. The network results in the LSTM, with CEC short term memories that last longer.
The key details are:
\begin{itemize}
 \item The $\circ$ operator is the Hadamard product, which is an element by element multiplication.
 \item The weights in the CEC are all fixed to the identity matrix and they are not trained.
 \item The output and the gating layer weights are also fixed to the identity matrix.
 \item It has been shown that the best results are obtained when initializing the feedback or forget gate, bias $\textbf{b}^{3}$, to all ones or larger values.
 \item Other weight and biases are randomly initialized to small numbers.
 \item The output of the gating layer generally connects to another layer or ML network with softmax transfer function.
 \item Multiple LSTM can be cascaded into each other.
\end{itemize}

We need to note that Deep Learning frameworks unroll these networks with delays and for each time step they create a physical layer and then use static backpropagation algorithm to calculate the gradients \cite{pascanu2013difficulty}. Then they roll the networks back and average the derivatives with respect to the weight and biases over the physical layers. The unrolling and rolling effect is only an approximation of the true gradient with respect to the weights.

\subsection{BERT}

\textit{BERT}, which stands for \textit{Bidirectional Encoder Representations from Transformers}, is a Language Model released by the Google AI Language team at the end of the year 2018 \cite{bert}. It has become the state-of-the-art model for many different \textit{Natural Language Undestanding} tasks, including sequence and document classification. This is best reflected by the fact that one can only see BERT-like models on the GLUE benchmark \href{https://gluebenchmark.com/leaderboard}{leaderboard}, with the sole exception of XLNet \cite{xlnet}, which then again is not so different from BERT. The success of BERT can be explained in part by its novel language modeling approach, but also by the use of the \textit{Transformer} \cite{transformer}, a Neural Network architecture based solely on \textit{Attention} mechanisms, which was introduced one year prior, replacing \textit{Recurrent Neural Networks} (RNNs) as the state-of-the-art Natural Language Understanding (NLU) techniques. We will give an overview of how Attention and Transformers work, and then explain BERT's architecture and its pre-training tasks.

\textit{Self-Attention}, the kind of Attention used on the Transformer, is essentially a mechanism that allows a Neural Network to learn representations of some text sequence influenced by all the words on the sequence. In a RNN context, this is achieved by using the hidden states of the previous tokens as inputs to the next time step. However, as the Transformer is purely feed-forward, it must find some other way of combining all the words together to map any kind of function in an a NLU task. It does so with the following equation

\begin{gather}
\text{Self-Attention} = \text{Softmax}\Bigg(\frac{QK^T}{\sqrt{d_k}}\Bigg)V
\end{gather}

\begin{figure}
    \centering
    \centerline{\includegraphics[width=\linewidth, height=2.3in]{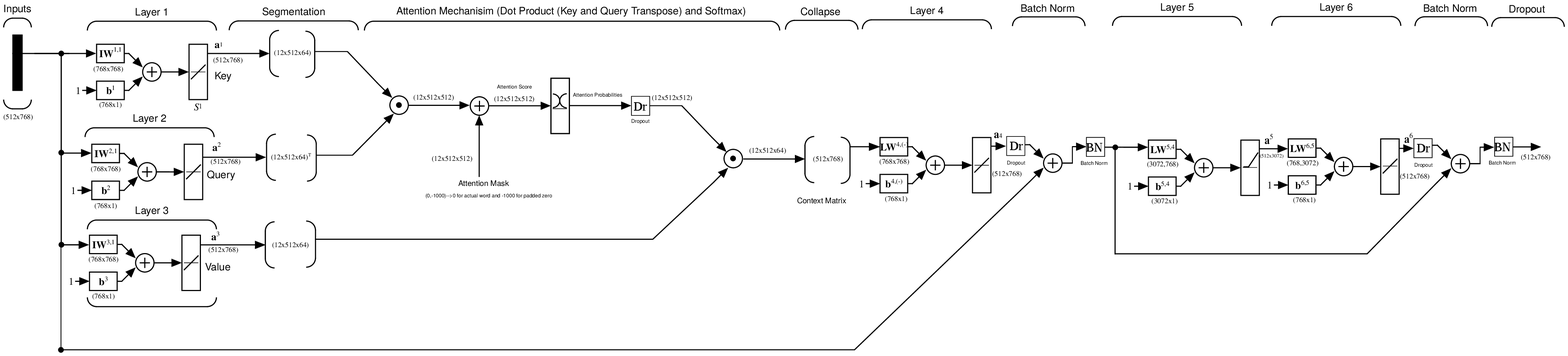}}
    \caption{One layer of base BERT}
    \label{BERT}
\end{figure}

Here, $Q$, $K$ and $V$ (\textit{query}, \textit{key} and \textit{value}) are matrices which are obtained by taking the dot product of some trainable weight matrices $W^Q$, $W^K$ and $W^V$ with the embedding matrix of our input sequence $X$. That is, $Q = W^Q X^T$, $K = W^K X^T$ and $V = W^V X^T$. Basically, each row on these matrices corresponds to one word, meaning that each word is mapped to three different projections of its embedding space. These projections serve as abstractions to compute the self-attention function for each word. The dot product between the query for word 1 and all the keys for words 1, 2, ..., $n$ tells us how ``similar'' each word is to word 1, a measure that is normalized by the \textit{softmax} function across all the words. The output of the softmax weights how much each word should contribute to the representation of the sequence that is drawn from word 1. Thus, the output of the self-attention transfer function for each word is a \textbf{weighted sum} of the values of all the words (including, and mainly, itself), by some parameters that are learnt to get the best representation that fits the problem at hand. $d_k$ is the dimension of the query vectors (512 for the Transformer, and 768 for base BERT and XLNet), and diving by its square root leads to more stable gradients.

The Transformer model goes one step further than simply computing a Self-Attention function, by implementing what is called \textit{Multi-Head Attention}. This is basically a set of $L$ Self-Attention computations, each on different sub-vectors, obtained from the original $Q$, $K$ and $V$ by breaking them up into $L$ different $Q_l$, $K_l$ and $V_l$ made up of $R/L$ components each, where $R$ is the embedding dimension (768 for the base BERT and XLNet) and $L$ the number of Attention Heads (12 for base BERT and XLNet). This is illustrated in Figure \ref{BERT} under ``Segmentation''. After the Self-Attention is computed for each $(Q_l, K_l, V_l)$, the original dimension is obtained by a simple concatenation (``Context Matrix'' in Figure \ref{BERT}). \\

\begin{figure}
\centering
\subfigure[Embedding Layer]          {\label{f2-a}   \includegraphics[width=3.2in, height=1.5in]{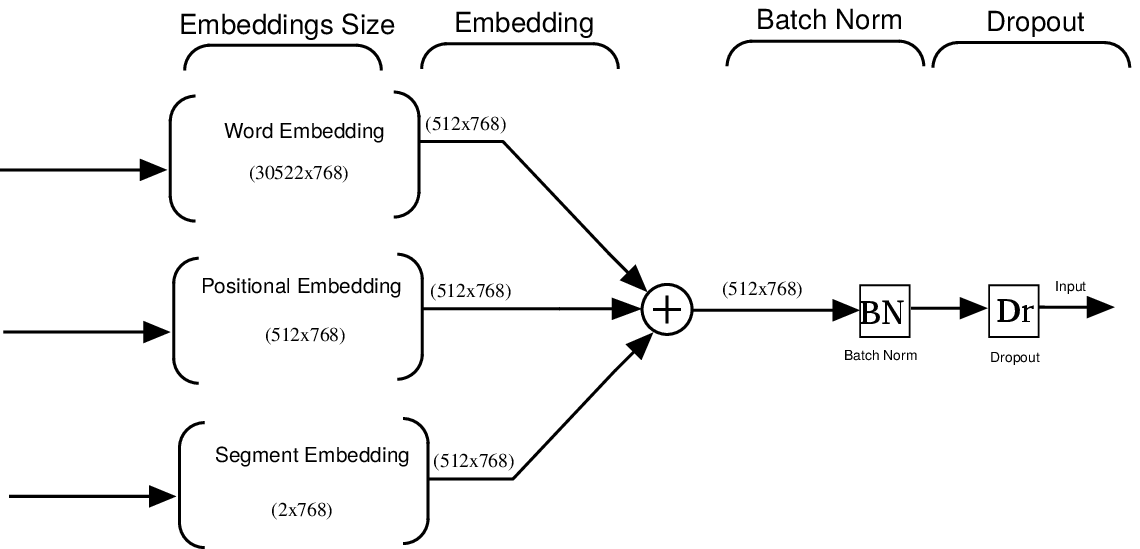}}
\subfigure[Pooler Layer]                 {\label{f2-b}   \includegraphics[width=2.8in, height=1.5in]{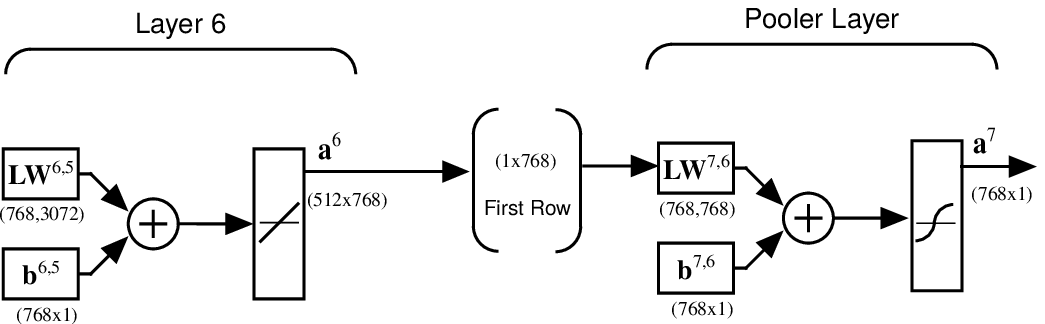}}
\label{f2}
\caption{BERT Embedding and Pooler Layer}
\end{figure}

Although up until this point we have only described the Encoder part of the Transformer, which is actually an Encoder-Decoder architecture, both BERT and XLNet use only an Encoder Transformer, so this is mainly all the architecture these Language Models are made of, with some key changes in the case of XLNet. Now we proceed to describe BERT's architecture from input to output, and also how it is pre-trained to learn a natural language. First, the actual words in the text are projected into an embedding dimension, which will be explained later in the context of Language Modeling. Once we have the embedding representation of each word, we input them into the first layer of BERT. Such layer, shown in Figure \ref{BERT}, consists mainly of a Multi-Head Attention Layer, which is identical to that of the Transformer, except for the fact that an \textit{attention mask} is added to the softmax input. This is done in order to avoid paying attention to padded 0s (which are necessary if one wants to do vectorized mini-batching). The attention mask is vector made up of 0s for the words we want the model to attend to (the actual words in the sequence), and of very small values (like -10,000) for the padded 0s. The sums of the keys and queries dot products with this mask will go into the softmax, making the attention scores for the masked padded 0s become practically 0. The output of this layer goes into a linear layer of size $R$x$R$, in order to learn a local linear combination of the Multi-Head Attention output. Batch Normalization is performed on the sum of the output of this layer (after a Dropout) and the input to the BERT layer. This is fed into yet another linear layer of size $R$x$R'$, where $R'=3072$ for the base BERT, followed by a \textit{GeLu} (Gaussian Error Linear Units) transfer function and another linear layer ($R'$x$R$) that maps the higher dimensions back to the embedding dimensions, with also Dropout and Batch Norm. This constitutes one BERT Layer, of which the base model has 12. The outputs of the first layer are treated as the hidden embeddings for each word, which the second layer takes as inputs and does the same kind of operations on them. Once we have gone through all the layers, the output for the first token (a special token ``[CLS]'' that remains the same for all input sequences) is passed onto another linear layer ($R$x$R$) with a \textit{tanh} transfer function. This layer (Figure \ref{f2-b}) acts as a pooler and its output is used as the representation of the whole sequence, which can finally allow learning multiple types of tasks by using other specific-purpose layers or even treating it as the sequence features to input into another kind of Machine Learning model. \\

\begin{figure}
    \centering
    \centerline{\includegraphics[width=\linewidth, height=3in]{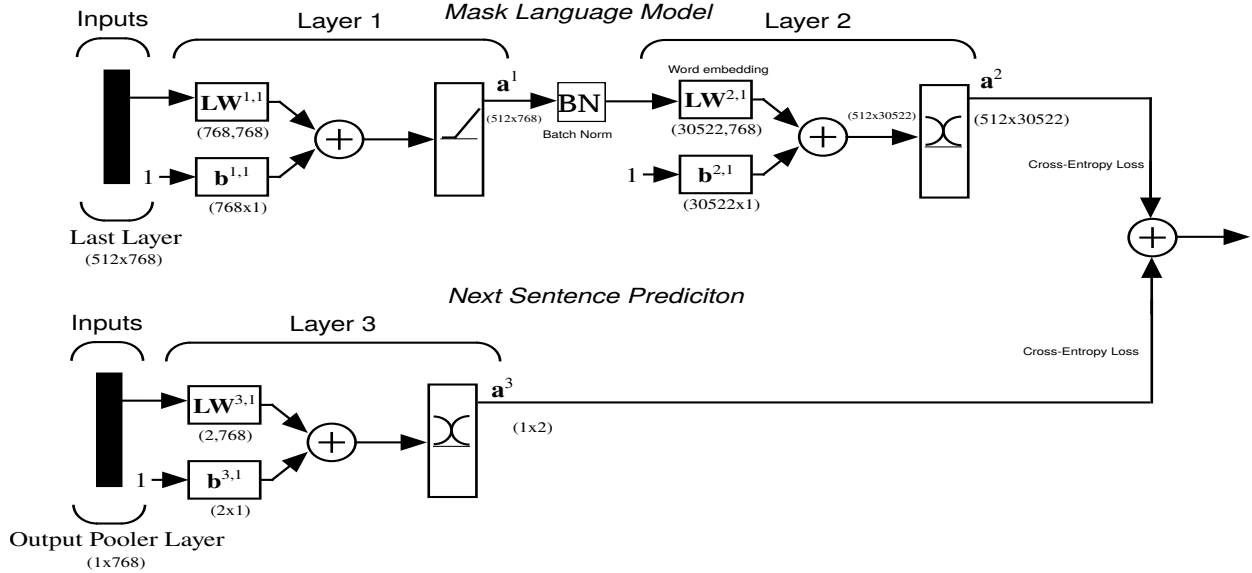}}
    \caption{BERT Pre-Training Heads}
    \label{BERTLM}
\end{figure}

Now that we have described BERT's architecture in detail, we will focus on the other main aspect that makes BERT so successful: BERT is, first and foremost, a Language Model. This means that the model is designed to learn useful knowledge about natural language from large amounts of unlabeled text, but also to retain and use this knowledge for supervised downstream tasks. The way Language Modeling usually works in a RNN scenario is just using the $n$ previous words as inputs to predict the next word $n+1$. The model cannot take as input the word $n+1$ or any words after it (although there are some bidirectional variants), so there is no need for special preprocessing of the text. However, as BERT is a feed-forward architecture that uses attention on all the words in some fixed-length sequence, if nothing is done, the model would be able to attend mainly to the very same word it is trying to predict. One solution would be cutting the attention on all the words after, and including, the target word. However, natural language is not so simple. More often than one would think, words within a sequence only make sense when taking the words after them as context. Thankfully, the attention mechanism can allow to capture both previous and future context, and one can stop the model from attending to the target word by \textit{masking} it (not to be confused with the attention mask used for the padded zeros). In particular, for each input sequence, 15 \% of the tokens are randomly masked, and then the model is trained to predict these tokens. The way this is done is taking the output of BERT, before the pooler, and mapping the vectors corresponding to each word to the vocabulary size with a linear layer, whose weights are the same as the ones from the input word embedding layer, although an additional bias is included, and then passing this to a softmax function in order to minimize a Categorical Cross-Entropy performance index that is computed with the predicted labels and the true labels (the ids on the token vocabulary, but only making the masked words contribute to the loss). Masking words is really straightforward: just replace them with the special token ``[MASK]''. This way, the network cannot use information from this word or any other masked words, aside from their position in the text. BERT was also pre-trained to predict whether a sentence B follows another sentence A (both randomly sampled from the text 50\% of the time, while the rest of the time sentence B is actually the sentence that comes after sentence A). Although recent research \cite{roberta} has shown that the same or even better results can be obtained without this second task, in the original implementation the model is optimized to minimize the sum of the losses from each task at the same time. The additional architecture just described is shown in Figure \ref{BERTLM}.

In addition to the usual word embeddings, positional embeddings are used to give the model information about the position of each word on the sequence (this is also done in the Transformer, although with some differences), and due to the next sentence prediction task and also for easy adaptation to downstream tasks such as question-answering, a segment embedding to represent each of the sentences is also utilized. The word embeddings used by BERT are \textit{WordPiece} embeddings \cite{wordpiece}, which consist in a tokenization technique in which the words are split into sub-word units. This helps handling \textit{out-of-vocabulary} words while keeping the actual vocabulary size small (30,522 unique word-pieces for BERT \textit{uncased}). The positional embeddings are look-up tables of size $512$x$R$, which assign a different embedding vector to each token based on its position within the sequence. $512$ was also chosen on the original Transformer as the maximum sequence length, mainly because Self-Attention's complexity is quadratic to the sequence length, due to the fact that it needs to compute the attentions of every word to every other word and also to themselves. While the first token in every input sequence will have the same positional embedding, the same applies to all the tokens belonging to the first sentence in the pair of sentences $A$ and $B$, i.e, the segment embeddings are look-up tables of size $2$x$R$. All of these embeddings have the same dimensions, so they can be simply added up element-wise to combine them together and obtain the input to the first Multi-Head Attention Layer, as shown in Figure \ref{f2-a}. Notice that these embeddings are learnable, so although pre-trained WordPiece are being used at the beginning for the word embeddings, these are being updated to represent the words in a better way during BERT's pre-training and fine-tuning tasks. This becomes even more crucial in the case of the positional and segment embeddings, which need to be learned from scratch. It is also worth noting that, although the embedding layers are technically look-up tables, which work with inputs of dimension $512$x$1$ (containing one unique token vocabulary id for each word), mathematically this is equivalent to having a linear layer (without bias) of size $512$x$R$, and one-hot-encoded inputs of dimension $512$xVocabularySize. The projection and weight update will be the same, but the first method is much faster because there is no matrix product involved, just a look-up (indexing) operation.


\subsection{XLNET}

XLNet is a language model introduced very recently \cite{xlnet} that makes use of the TransformerXL \cite{transformerXL} to incorporate information from previous sequence/s in order to process the current sequence, achieving a regressive effect at the sequence level. To do so, it employs a \textit{relative positional encoding} and a permutation language modeling approach. Although BERT and XLNet share a lot of similarities, there are some key differences that need to be explained. \\

\begin{figure}
    \centering
    \centerline{\includegraphics[width=\linewidth, height=4.5in]{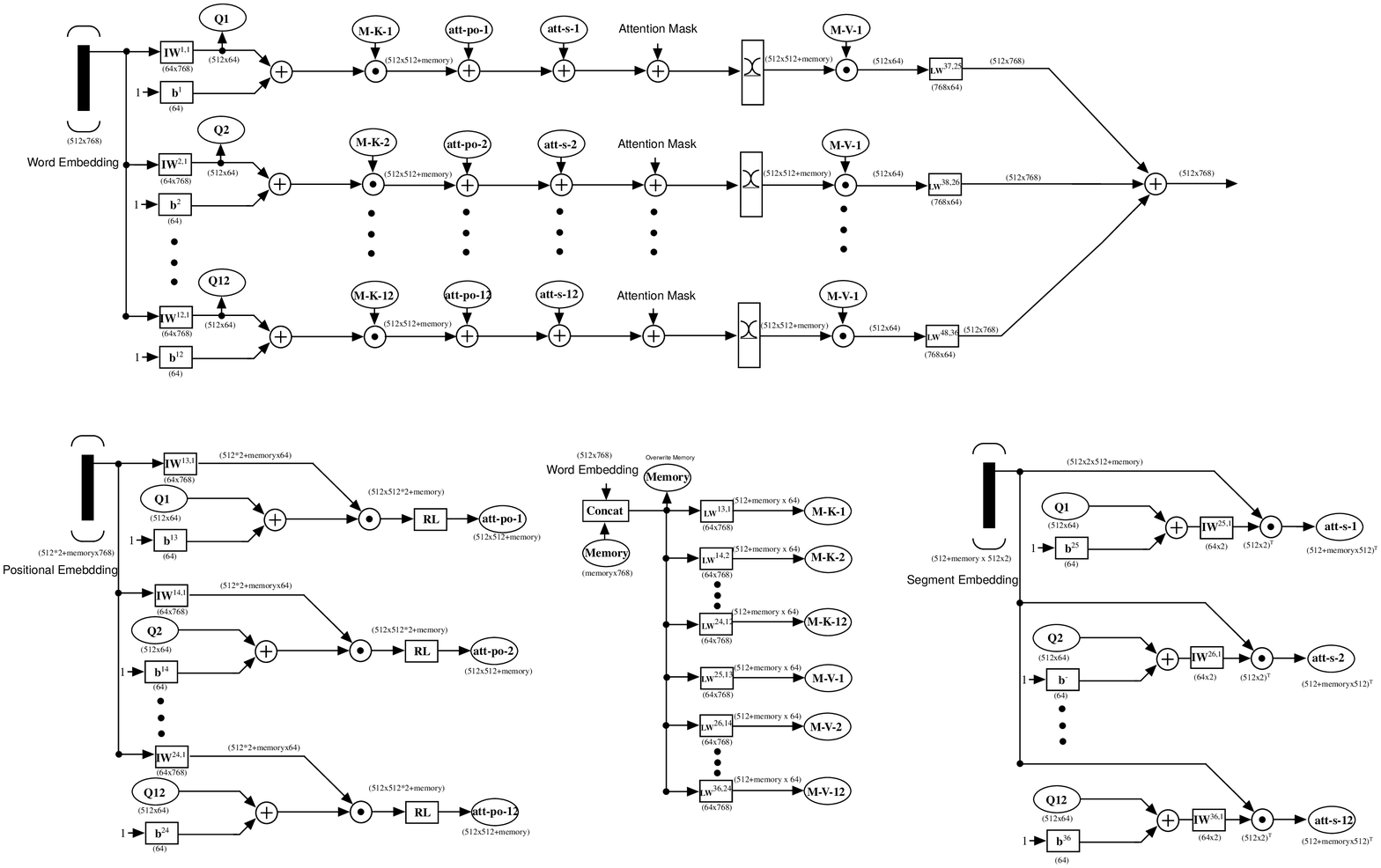}}
    \caption{One layer of base XLNet (content stream)}
    \label{XLNet}
\end{figure}

Firstly, XLNet's Multi-Head Attention's core operation is different than the one implemented in BERT and in the Transformer. In this case, instead of just breaking up the original $Q$, $K$ and $V$ into $L$ different $Q_l$, $K_l$ and $V_l$; $L$ linear layers (for each) are used to map the input to the Multi-Head Attention layer into these different $Q_l$, $K_l$ and $V_l$, and thus no intermediate $Q$, $K$ and $V$ are computed. This results in the three linear layers of $R$x$R$ being replaced by $3L$ linear layers of $R/L$ x $R$, which map the input into smaller subspaces (with the same number of dimensions which add up to the original dimension). These several (and parallel) computations on different dimensions produce more variability, allowing each word to attend more to other words and not only to itself, which results in a final richer representation of each word, calculated by adding up the results of mapping back each of the sub-representations to the original embedding dimension $R$ with again 12 linear layers. This is expressed with the following equation, where $X$ is the input. Note that the actual implementation is a bit more complex, as shown in Figure \ref{XLNet}, but the heart of the operation is indeed in equation (\ref{XLNETMH}).

\begin{gather} \label{XLNETMH}
\text{MultiHead}(X) = \sum_{l=1}^{12} \Bigg [ \text{Softmax}\Bigg(\frac{\overbrace{XW^Q_l}^{Q_l} (\overbrace{X W^K_l}^{K_l})^T}{\sqrt{d_k}}\Bigg)\overbrace{X W^V_l}^{V_l} \Bigg ] W_l^O
\end{gather}

\begin{figure}
    \centering
    \centerline{\includegraphics[width=\linewidth, height=2in]{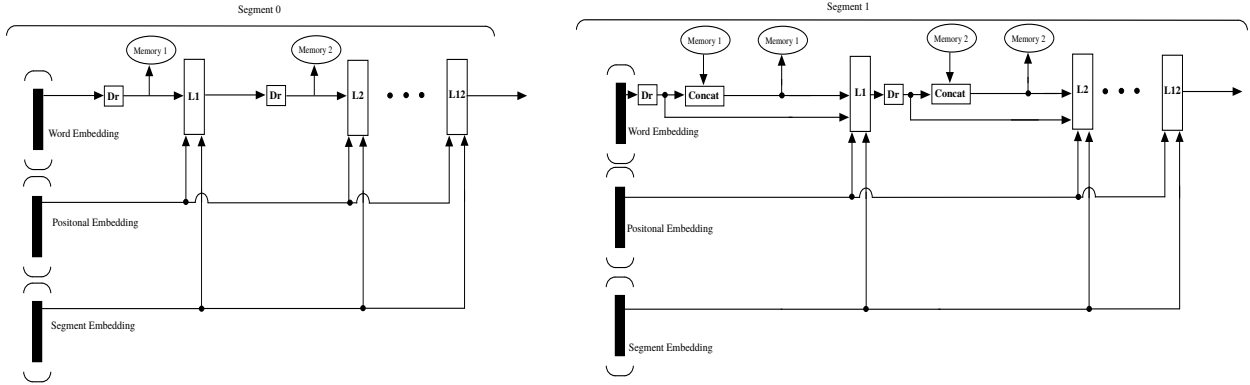}}
    \caption{12 Layer XLNet (content stream)}
    \label{XLNetAR}
\end{figure}

Secondly, apart from this, XLNet's Attention is different from BERT's in two ways: 1. The keys and values (but not the queries) of the current sequence and for each layer depend on the hidden states of the previous sequence/s, based on a memory length hyper-parameter. That is, let the hidden state (output) of layer $m$ for the previous sequence be a matrix $h^m_{t-1}$ of dimensions $512$ x $768$, then if we choose a memory length of $\text{mem}_{\text{len}} = 10$ tokens, the key and value of the current sequence for layer $m+1$ will be computed by concatenating to $h^{m}_t$ the last ten vectors of $h^m_{t-1}$ and then projecting the result using the $W^{K, m+1}_l$ and $W^{V, m+1}_l$ matrices. This recurrence mechanism at the sequence level is illustrated in Figure \ref{XLNetAR}. If the memory length is greater than 512, we can even reuse information from the two last sequences, although this becomes quadratically expensive. 2. The operation just described is only applied to the word embeddings, and not to the sum of the three kinds of embeddings ($h^m$ are just the word embeddings). The other two are used in a different way. The (relative) positional embeddings (encodings is a more suitable name) are computed by the following equation

\begin{gather}
p_{\text{pos}} = \text{concat} \Bigg[ \sin \Big(e_{\text{inv-inds}} \otimes  p_{\text{inds}} \Big), \cos \Big(e_{\text{inv-inds}} \otimes  p_{\text{inds}} \Big) \Bigg]
\end{gather}

where $p_{\text{inds}} = [\text{mem}_{\text{len}}+\text{seq}_{\text{len}}, \text{mem}_{\text{len}} +\text{seq}_{\text{len}}-1, \text{mem}_{\text{len}} + \text{seq}_{\text{len}}-2, ..., 0, -1, -2, ..., -\text{seq}_{\text{len}}+1]$ and $e_{\text{inv-inds}_i} = \Big (10000^{e_{\text{inds}_i}/784}\Big)^{-1}$, with $e_{\text{inds}} = [0, 2, 4, ..., 766]$. Note that these are also different from BERT's in the sense that they are not being learnt. $p_{\text{pos}}$, of shape $(2\text{seq}_{\text{len}} + \text{mem}_{\text{len}})$ x $768$, is projected into $L$ positional keys $K_{l,p}$ by learnable matrices $W^{K, p}_l$. The dot products between these and $L$ positional queries obtained from the original queries by adding them up with learnable biases $b^{Q,p}_l$ are performed, and then the second to the $\text{seq}_{\text{len}} + 1$ elements are obtained from the memory dimension after performing a relative shift between this dimension and the current sequence dimension, resulting in positional attention scores which are added up to the regular attention scores before going into the softmax. This way, XLNet can perform a smarter attention to both the words on the previous sequence/s and the current sequence, by using this information that is being learnt based on the relative position of each word with respect to each other word of each sequence. To distinguish between current and previous sequence/s, a segment embedding is also utilized, which consists simply of a one-hot-encoded matrix of $(\text{seq}_{\text{len}}+\text{mem}_{\text{len}})$x$\text{seq}_{\text{len}}$x$2$: we have a $1$ if word $i$ and word $j$ belong to the same sequence, and $0$ otherwise. Before attending to this segment encoding (which acts as a unique segment key), the original queries are again added up with biases $b^{Q,s}_l$ and then projected by $L$ weight matrices into $L$ $Q_{l, s}$, of shape $512$x$2$. The result is also added to the attention scores before the softmax, and the operation described on equation (\ref{XLNETMH}) with the values is performed. The rest is almost identical to BERT (layers 5 and 6 on Figure \ref{BERT}, layer 4 and the batch norm after it are omitted), taking into account that the output of the first XLNet layer acts as hidden word embeddings that go into to the second layer, while the positional and segment encodings inputed to this layer remain the same as the ones inputed to the first layer, as shown in Figure \ref{XLNetAR}. The detailed architecture is shown in Figure \ref{XLNet}. \\

While the architecture differences have been listed above, XLNet also differs from BERT in their pre-training tasks. XLNet is pre-trained by a permutation language modeling approach. This means that, for any sequence, there are \text{sequence}\_\text{length}! permutations of the factorization order, and an AR language modeling can be performed by maximizing the likelihood under the forward autoregressive factorization. Note that the order of the sequence remains unchanged: the permutation only affects which words are attended to, by changing the attention mask before the softmax: to predict word $k$, the attention mask is set to very small numbers for words with $i>k$, so that only the words before and including $k$ on the current factorization order are used to compute the attention. The trick here is that the words that come before $k$ change with each permutation, but their positions are kept constant within the sequence, allowing XLNet to capture bidirectional context. Additionally, due to the fact that utilizing permutations causes slow convergence, XLNet is pre-trained to predict only the last 16.67 \% of tokens in each factorization.

In order to use the position of the token $k$ that is going to be predicted, but not its content, XLNet introduces a new type of query. The same kind of Multi-Head attention is performed, starting from a randomly initialized vector (or vectors if we are predicting more than one token at the same time). This vector is projected by the same $L$ linear layers as the normal query to obtain the new type of query, which attends to the same keys and values as the regular query, but with the new attention mask explained before (with the difference that the element corresponding to word $k$ is also set to a very small value). So basically, in the pre-training task, this new Multi-Head Attention (named query stream) and the one from Figure \ref{XLNet} (named content stream) are performed at the same time layer by layer, because the query stream needs the outputs of each layer from the content stream to get the content keys and the values to perform the attention on the next layer. The content stream can see the content of the words that come before $k$ in the factorization order, and also $k$, while the query stream can only see the content of the words that come before $k$. After going through the 12 XLNet layers and projecting the output of this new query with a linear layer of VocabularySize x target\_tokens, the Cross-Entropy loss with the indexes of the real tokens is computed and the model's parameters are updated to minimize this loss. The method just described allows XLNet to be pre-trained without the need to replace the target tokens with the special token ``[MASK]'', which is not present at all during fine-tuning.


\section{Fine Tuning and Experiments}

In this section we provide an overview of how neural language model fine-tuning is done for a downstream classification task such as essay scoring, as well as explain the experiments we did in order to improve performance. The output layer/s that were used for the pre-training task/s are replaced with a single classification layer. This layer has the same number of neurons as labels (possible scores for the essays), with a \textit{softmax} activation function, which is then used, together with the target, to compute a \textit{cross-entropy} performance index as a loss function. In the case of BERT, the last hidden state of the first (and special) token ``[CLS]'' is used as the representation of the whole essay. Because this representation needs to be adjusted to the particular problem at hand, the whole model is trained. This differs from the way in which transfer learning is done on images, where, if the model was pretrained using at least some images similar to the task at hand, updating all the parameters does not usually provide a boost in performance that is justifiably by the much longer training time. Regarding XLNet, the same method is applied but now the ``[CLS]'' token is located at the end of the essay.

In theory, the model should retain most of the knowledge it learnt about the English language during the pre-training tasks. This would provide not only a much better initialization, which drastically reduces the downstream training time, but also an increase in performance when compared with other Neural Networks that need to learn natural language from random initial conditions from a much smaller corpus. However, in practice, various problems can arise such as \textit{catastrophic forgetting}, which means the model forgets very quickly what it had learnt previously, rendering the main point of transfer learning almost useless. There are various ways of dealing with this: we try \textit{gradual unfreezing}, \textit{discriminative fine-tuning} and a combination of both as proposed in \cite{ulmfit}. Gradual unfreezing consists of only training the last layer on the first epoch, which contains the least general information about the language, and then unfreezing one more layer per epoch, from last to first. On the other hand, discriminative fine-tuning consists on using different learning rates for different layers, as they capture different kinds of features on Deep Networks \cite{deepfeat}. In particular, BERT has been shown to attend to different kinds of words and capture diverse linguistic notions on different attention heads \cite{bertat}. The learning rate across layers $m$ follows the formula $\alpha^m = \xi \alpha^{m+1}$, where $\xi$ is a decay factor usually set close to 1 \cite{bertclass}. Another closely related problem is \textit{overfitting}. To mitigate this, we try using the model's hidden states at different layers and also some data-preprocessing to force the model to focus more on other kinds of words, as well as different dropout values. Lastly, we also use an ensemble of different models trained with the different approaches.

We run our experiments using \texttt{pytorch-transformers} implementations of BERT and XLNet. We choose Adam as the optimizer, as in the original papers, and try different learning rates, narrowing the best values to either $e^{-5}$ or $5e^{-6}$. We also try different \textit{warmup schedules}, and find that they make no significant difference. Regarding BERT, there are currently two main versions: ``cased'' and ``uncased''. We find that overall ``uncased'' works slightly better, although for some items the ``cased'' version is superior. However, the difference is still very small. For XLNet, the only available version is ``cased''. We also compare the base and large versions and find that they perform very similarly, so using the large versions is not worth it, given that they are much more expensive to fine-tune. Thus, all the results shown are for the base versions. The same applies to the batch size, so we end up using the largest we could fit in a 12GB GPU, i.e, 9 for BERT and 8 for XLNet.

Due to the fact that BERT and XLNet were pre-trained with sequences of 512 tokens (510 when taking into account ``[CLS]'' and ``[SEP]''), and some of our essays are quite longer than that, we use a sliding-window approach in which longer essays are split into two or more sequences of 510 tokens. We force an overlapping of the last of these sequences with the second-to-last, in order to avoid meaningless padding on the last split. For prediction, we just round the average of the scores on each of these splits. Although we also experimented imputing only the first 510 tokens, or the first 128 and last 382, as proposed in \cite{bertclass}, this did not make any significant difference, and even if it did, it should be avoided because in the context of essay scoring it could be argued to be unethical.


\section{Results}

In this section we evaluate more in-depth each of the things we tried on the development set, and then provide the results on the test set by picking the best model on the development set. Table \ref{bcompresults} shows the dev \textit{qwk} percentage difference between each combination and the base try, which is just using BERT/XLNet as they are. It can be seen that, overall, the methods to avoid catastrophic forgetting do not work very well, although for particular items they can give a small boost in performance. Their combination (1+2) also performs poorly on all the items except for BERT on item 8. Increasing the dropout probability is neither a good idea, and when it helps, it is only slightly. However, in the case of BERT, decreasing the input complexity (removing stop-words), the model complexity (using only the three first layers) and a combination of both seems to actually reduce overfitting and works the best overall, which is good news because it is much more inexpensive than running a combination of the previous methods, even more so when trying different learning rates and warm-up schedules for each of them. On the other hand, XLNet only sees an increase in performance for two items when removing stop-words, and using three layers does not help either. These findings suggests that BERT is more flexible than XLNet, or at least that it can adapt better to extreme changes in the architecture and input levels. Regarding catastrophic forgetting, it looks like XLNet does witness more improvement than BERT for the items in which either gradual unfreezing, discriminative finetuning or their combination boost performance.

\begin{table}
\renewcommand{\arraystretch}{1.2}
\caption{BERT and XLNet Results of various experiments for each item}
\resizebox{\textwidth}{!}{%
\begin{tabular}{|l|l|l|l|l|l|l|l|l|l|}
\hline
BERT Experiments / Item                    & $\Delta$ 1 (\%) & $\Delta$ 2 (\%) & $\Delta$ 3 (\%) & $\Delta$ 4 (\%) & $\Delta$ 5 (\%) & $\Delta$ 6 (\%) & $\Delta$ 7 (\%) & $\Delta$ 8 (\%) & Mean $\Delta$ (\%)\\ \hline
(1) Gradual Unfreezing                     & -4.58          & -5.17          & \textcolor{darkgreen}{+0.04}          & -1.49          & -5.76          & -4.19  & -18.87         & -6.41          & -5.80 \\ \hline
(2) Discriminative Finetuning ($\xi=0.95$) & -3.59          & -0.04          & \textcolor{darkgreen}{+0.70}          & \textbf{+1.17} & -1.61          & -1.08  & -0.14          & -10.20         & -1.85 \\ \hline
1+2                                        & -2.69          & -3.71          & -0.54          & -3.58          & -6.07          & -7.85  & -10.8          & \textbf{+1.38} & -4.23 \\ \hline
(3) Dropout (0.2)                          & -3.77          & \textcolor{darkgreen}{+0.49}          & \textcolor{darkgreen}{+0.93}          & -0.12          & -2.11          & -0.54  & -3.57          & -31.83         & -5.07 \\ \hline
1+3                                        & -5.29          & -7.29          & \textcolor{darkgreen}{+0.32}          & -0.99          & -4.27          & -2.10  & -8.49          & -7.18          & -4.41 \\ \hline
2+3                                        & \textcolor{darkgreen}{+0.57}          & \textcolor{darkgreen}{+0.43}          & \textcolor{darkgreen}{+0.23}          & -0.45          & -3.47          & -0.42  & -8.22          & -5.00          & -2.04 \\ \hline
1+2+3                                      & -3.65          & -7.72          & -0.36          & -2.13          & -7.15          & -13.52 & -19.06         & -17.93         & -8.94 \\ \hline
(4) Remove Stop-Words                      & \textbf{+2.60} & -2.43          & -0.66          & -0.92          & -4.03          & -2.18  & \textcolor{darkgreen}{+0.69}          & -0.17          & -0.89 \\ \hline
(5) 3 Layers                               & -0.41          & \textcolor{darkgreen}{+0.23}          & -0.55          & -0.52          & \textbf{+0.12} & -1.14  & \textcolor{darkgreen}{+0.37}          & -2.71          & -0.58 \\ \hline
4 + 5                                      & \textcolor{darkgreen}{+1.82}          & \textbf{+1.35} & \textbf{+2.04} & -0.05          & -1.90          & -2.20  & \textbf{+0.77} & -0.65          & \textbf{+0.15} \\ \hline
\end{tabular}
}

\vspace{2.5mm}

\resizebox{\textwidth}{!}{%
\begin{tabular}{|l|l|l|l|l|l|l|l|l|l|}
\hline
XLNet Experiments / Item                                   & $\Delta$ 1 (\%) & $\Delta$ 2 (\%) & $\Delta$ 3 (\%) & $\Delta$ 4 (\%) & $\Delta$ 5 (\%) & $\Delta$ 6 (\%) & $\Delta$ 7 (\%) & $\Delta$ 8 (\%) & Avg $\Delta$ (\%)\\ \hline
(1) Gradual Unfreezing                     & -4.46          & -6.00  & -2.86  & \textbf{+1.38} & -3.31          & -1.34  & -1.95          & -0.26          & -2.35 \\ \hline
(2) Discriminative Finetuning ($\xi=0.95$) & \textcolor{darkgreen}{+2.41}          & -2.02  & -1.05  & -1.02          & -1.75          & -0.41  & -1.26          & \textbf{+4.50} & -0.08 \\ \hline
1+2                                        & -4.58          & -6.38  & -1.28  & -1.93          & -3.47          & -2.46  & -0.83          & -5.51          & -3.31 \\ \hline
(3) Dropout (0.2)                          & -0.59          & -2.28  & -3.16  & -2.65          & -2.53          & -6.88  & -6.89          & -1.87          & -3.36 \\ \hline
1+3                                        & -7.57          & -19.41 & -7.91  & -9.56          & -5.61          & -10.29 & -3.42          & -9.07          & -9.11 \\ \hline
2+3                                        & -3.97          & -6.80  & -1.04  & -1.02          & -2.21          & -4.60  & -5.51          & -18.97         & -5.52 \\ \hline
1+2+3                                      & -8.89          & -23.34 & -11.09 & -5.89          & -7.31          & -16.66 & -16.32         & -37.86         & -15.92 \\ \hline
(4) Remove Stop-Words                      & \textbf{+2.66} & -4.19  & -1.72  & -0.79          & -1.59          & -2.36  & \textbf{+2.04} & -3.19          & -1.14 \\ \hline
(5) 3 Layers                               & -0.96          & -0.29  & -4.07  & -1.54          & \textbf{+0.06} & -3.70  & -7.95          & -20.06         & -4.81 \\ \hline
4 + 5                                      & \textcolor{darkgreen}{+0.94}          & -1.26  & -2.49  & -2.89          & -0.12          & -1.32  & -8.58          & -20.44         & -4.52 \\ \hline
\end{tabular}
}
\label{bcompresults}
\end{table}

Table \ref{crossvalresults} shows the final results on each item for BERT, XLNet, a BERT ensemble, an XLNet ensemble and a BERT + XLNet ensemble. The first two ensembles consist of 6 models obtained using the different experiments from above. We tried taking a majority vote (using the best model out of these 6 to decide when there is a tie), and rounding the mean of the scores predicted by each model. Both methods performed similarly on items 1 to 6, but the majority vote performed significantly poorer on items 7 and 8. The BERT + XLNet ensemble consists of 12 models, i.e, it combines the models from the two other ensembles together. We also show the results for the LSTM from \cite{lstm} and their ensemble, which consists on 10 LSTMs and 10 LSTMs with a convolutional layer before them, and which also arrives at the final prediction by taking the mean of the scores. The last two rows correspond to the Bag of Words model and the inter-human agreement.

\begin{table}
\renewcommand{\arraystretch}{1.2}
\caption{qwks for the different items}
\resizebox{\textwidth}{!}{%
\begin{tabular}{|l|l|l|l|l|l|l|l|l|l|}
\hline
Item                             & 1 qwk (\%)     & 2 qwk (\%)     & 3 qwk (\%)     & 4 qwk (\%)     & 5 qwk (\%)     & 6 qwk (\%)     & 7 qwk (\%)     & 8 qwk (\%)    & Avg qwk(\%) \\ \hline
BERT                             & 79.20          & 67.99          & \textbf{71.52} & 80.08          & 80.59          & 80.53          & 78.51          & 59.58          &  74.75      \\ \hline
XLNet                            & 77.69          & 68.06          & 69.29          & 80.62          & 78.33          & 79.37          & 78.67          & 62.68          &  74.34      \\ \hline
LSTM \cite{lstm}                 & 77.50          & 68.70          & 68.30          & 79.50          & \textbf{81.80} & 81.30          & 80.50          & 59.40          &  74.63      \\ \hline
BERT Ensemble                    & 80.21          & 67.21          & 70.82          & 81.56          & 80.63          & 81.47          & 80.42          & 59.74          &  75.26      \\ \hline
XLNet Ensemble                   & 80.49          & 68.59          & 70.09          & 79.56          & 79.94          & 80.54          & 80.02          & 59.76          &  74.87      \\ \hline
BERT + XLNet Ensemble            & 80.78          & \textbf{69.67} & 70.31          & \textbf{81.90} & 80.82          & 81.45          & 80.67          & 60.46          &  75.76      \\ \hline
LSTM (+CNN) Ensemble \cite{lstm} & \textbf{82.10} & 68.80          & 69.40          & 80.50          & 80.70          & \textbf{81.90} & \textbf{80.80} & \textbf{64.40} &  \textbf{76.08}      \\ \hline
EASE (Bag of Words) \cite{lstm}                 & 78.10          & 62.10          & 63.00          & 74.90          & 78.20          & 77.10          & 72.70          & 53.40          &  69.90      \\ \hline
H1-H2 Agreement                  & 72.08          & 81.23          & 76.90          & 85.10          & 75.27          & 77.59          & 72.09          & 62.03          &  75.29      \\ \hline
\end{tabular}
}
\label{crossvalresults}

\end{table}

Regarding the individual models, BERT, XLNet and the LSTM obtain very similar average qwk across all the items. This suggests that the essay scoring problem has reached its ceiling in terms of modeling, at least for now. When compared to their individual versions, the ensemble boost performance by 0.51 \% for BERT, 0.53 \% for XLNet, and 1.45 \% for the LSTM. The main difference between the Language Models ensembles and the LSTM ensemble that may account for an almost 3x bigger delta in favor of the LSTM is the amount of models (6 vs 20), although it is possible that the convolution layer on the LSTM produces more variability than using different number of layers and altering the inputs to the Language Models. The BERT + XLNet ensemble (12 models) does better (+1.01 \% from BERT and +1.42 \% from XLNet), which points to the first reason being more likely.

When compared with the Bag of Words method, Neural Networks show a significant superiority, with a 6 \% higher qwk on average. What is more, there is no single item for which the Bag of Words performs better. And although individual networks are still a bit below the inter-human agreement, the ensemble of these models are actually beating humans by 0.79 \% in the case of the LSTM and by 0.47 \% in the case of BERT + XLNet on average. Item by item, Neural Networks achieve higher-than-human qwk on 5 out of 8.


\section{Conclusions}

Transfer learning and language models enhanced the performance of analyzing texts in natural language processing. In this paper, we demonstrated the two major transformer based neural network models which improved the result of essay scoring on the Kaggle dataset. BERT and XLNet are discussed in a very detailed manner to researchers for further improvements. The results of BERT and XLNet are compared with other traditional methods and human standards. Overall, we got better results that human and rule based techniques. Our major contribution is explaining the network architectures and generalizing it with simple notation, and implementing a classification technique using these models on the essay scoring problem to get an automated engine. This engine tends to be more reliable than humans and save a lot of time and money for grading essays in a large scale.

\section{Acknowledgements}

I would like to thank Balaji Kodeswaran, and Paul van Wamelen for their support and discussions.

\newpage
\bibliographystyle{IEEEtran}
\bibliography{mybib}

\begin{thebibliography}{10}
\providecommand{\url}[1]{#1}
\csname url@samestyle\endcsname
\providecommand{\newblock}{\relax}
\providecommand{\bibinfo}[2]{#2}
\providecommand{\BIBentrySTDinterwordspacing}{\spaceskip=0pt\relax}
\providecommand{\BIBentryALTinterwordstretchfactor}{4}
\providecommand{\BIBentryALTinterwordspacing}{\spaceskip=\fontdimen2\font plus
\BIBentryALTinterwordstretchfactor\fontdimen3\font minus
  \fontdimen4\font\relax}
\providecommand{\BIBforeignlanguage}[2]{{%
\expandafter\ifx\csname l@#1\endcsname\relax
\typeout{** WARNING: IEEEtran.bst: No hyphenation pattern has been}%
\typeout{** loaded for the language `#1'. Using the pattern for}%
\typeout{** the default language instead.}%
\else
\language=\csname l@#1\endcsname
\fi
#2}}
\providecommand{\BIBdecl}{\relax}
\BIBdecl

\bibitem{Page1967}
E.~B. Page, ``Grading essays by computer: progress report.'' in
  \emph{Proceedings of the Invitational Conference on Testing Problems}, 1967,
  pp. 87--100.

\bibitem{Page1968}
------, ``The use of the computer in analyzing student essays.'' in
  \emph{International Review of Education}, 1968, pp. 210--225.

\bibitem{IRR}
K.~L. Gwet, \emph{Handbook of inter-rater reliability: The definitive guide to
  measuring the extent of agreement among raters}.\hskip 1em plus 0.5em minus
  0.4em\relax Advanced Analytics, LLC, 2014.

\bibitem{alikaniotis2016automatic}
D.~Alikaniotis, H.~Yannakoudakis, and M.~Rei, ``Automatic text scoring using
  neural networks,'' \emph{arXiv preprint arXiv:1606.04289}, 2016.

\bibitem{sakaguchi2015effective}
K.~Sakaguchi, M.~Heilman, and N.~Madnani, ``Effective feature integration for
  automated short answer scoring,'' in \emph{Proceedings of the 2015 conference
  of the North American Chapter of the association for computational
  linguistics: Human language technologies}, 2015, pp. 1049--1054.

\bibitem{Shermis2015}
\BIBentryALTinterwordspacing
M.~D. Shermis, ``Contrasting state-of-the-art in the machine scoring of
  short-form constructed responses,'' \emph{Educational Assessment}, vol.~20,
  no.~1, pp. 46--65, 2015. [Online]. Available:
  \url{https://doi.org/10.1080/10627197.2015.997617}
\BIBentrySTDinterwordspacing

\bibitem{yannakoudakis2015evaluating}
H.~Yannakoudakis and R.~Cummins, ``Evaluating the performance of automated text
  scoring systems,'' in \emph{Proceedings of the Tenth Workshop on Innovative
  Use of NLP for Building Educational Applications}, 2015, pp. 213--223.

\bibitem{hagan1994training}
M.~T. Hagan and M.~B. Menhaj, ``Training feedforward networks with the
  marquardt algorithm,'' \emph{IEEE transactions on Neural Networks}, vol.~5,
  no.~6, pp. 989--993, 1994.

\bibitem{imagenet}
A.~Krizhevsky, I.~Sutskever, and G.~E. Hinton, ``Imagenet classification with
  deep convolutional neural networks,'' in \emph{Advances in neural information
  processing systems}, 2012, pp. 1097--1105.

\bibitem{mikolov2013efficient}
T.~Mikolov, K.~Chen, G.~Corrado, and J.~Dean, ``Efficient estimation of word
  representations in vector space,'' \emph{arXiv preprint arXiv:1301.3781},
  2013.

\bibitem{JAFARI2018312}
\BIBentryALTinterwordspacing
A.~H. Jafari and M.~T. Hagan, ``Application of new training methods for neural
  model reference control,'' \emph{Engineering Applications of Artificial
  Intelligence}, vol.~74, pp. 312 -- 321, 2018. [Online]. Available:
  \url{http://www.sciencedirect.com/science/article/pii/S0952197618301490}
\BIBentrySTDinterwordspacing

\bibitem{bahdanau2014neural}
D.~Bahdanau, K.~Cho, and Y.~Bengio, ``Neural machine translation by jointly
  learning to align and translate,'' \emph{arXiv preprint arXiv:1409.0473},
  2014.

\bibitem{lstm}
K.~Taghipour and H.~T. Ng, ``A neural approach to automated essay scoring,''
  \emph{Conference on Empirical Methods in Natural Language Processing}, no.
  1905.05583, p. 1882–1891, November 2016.

\bibitem{opengpt2}
A.~Radford, J.~Wu, R.~Child, D.~Luan, D.~Amodei, and I.~Sutskever, ``Language
  models are unsupervised multitask learners,'' \emph{OpenAI Blog}, vol.~1,
  no.~8, 2019.

\bibitem{bert}
J.~Devlin, M.-W. Chang, K.~Lee, and K.~Toutanova, ``Bert: Pre-training of deep
  bidirectional transformers for language understanding,'' \emph{arXiv}, no.
  1810.04805, Oct 2018.

\bibitem{xlnet}
Z.~Yang, Z.~Dai, Y.~Yang, J.~Carbonel, R.~Salakhutdinov, and Q.~V. Le, ``Xlnet:
  Generalized autoregressive pretraining for language understanding,''
  \emph{arXiv}, no. 1906.08237, Jun 2019.

\bibitem{elmo}
M.~E. Peters, M.~Neumann, M.~Iyyer, M.~Gardner, C.~Clark, K.~Lee, and
  L.~Zettlemoyer, ``Deep contextualized word representations,'' in \emph{Proc.
  of NAACL}, 2018.

\bibitem{williamson2012framework}
D.~M. Williamson, X.~Xi, and F.~J. Breyer, ``A framework for evaluation and use
  of automated scoring,'' \emph{Educational measurement: issues and practice},
  vol.~31, no.~1, pp. 2--13, 2012.

\bibitem{rubrics}
J.~Arter, ``Rubrics, scoring guides, and performance criteria: Classroom tools
  for assessing and improving student learning.'' 2000.

\bibitem{cohen1960coefficient}
J.~Cohen, ``A coefficient of agreement for nominal scales,'' \emph{Educational
  and psychological measurement}, vol.~20, no.~1, pp. 37--46, 1960.

\bibitem{PEG}
E.~B. Page, ``The imminence of... grading essays by computer,'' \emph{The Phi
  Delta Kappan}, vol.~47, no.~5, pp. 238--243, 1966.

\bibitem{embedding}
T.~Mikolov, I.~Sutskever, K.~Chen, G.~S. Corrado, and J.~Dean, ``Distributed
  representations of words and phrases and their compositionality,'' in
  \emph{Advances in neural information processing systems}, 2013, pp.
  3111--3119.

\bibitem{hochreiter1997long}
S.~Hochreiter and J.~Schmidhuber, ``Long short-term memory,'' \emph{Neural
  computation}, vol.~9, no.~8, pp. 1735--1780, 1997.

\bibitem{NND}
M.~T. Hagan, H.~B. Demuth, M.~H. Beale, and O.~D. Jes\'us, \emph{Neural Network
  Design, 2nd Edition}.\hskip 1em plus 0.5em minus 0.4em\relax PWS Publishing,
  Boston.

\bibitem{chung2015gated}
J.~Chung, C.~Gulcehre, K.~Cho, and Y.~Bengio, ``Gated feedback recurrent neural
  networks,'' in \emph{International Conference on Machine Learning}, 2015, pp.
  2067--2075.

\bibitem{jafari2015enhanced}
A.~H. Jafari and M.~T. Hagan, ``Enhanced recurrent network training,'' in
  \emph{2015 International Joint Conference on Neural Networks (IJCNN)}.\hskip
  1em plus 0.5em minus 0.4em\relax IEEE, 2015, pp. 1--8.

\bibitem{pascanu2013difficulty}
R.~Pascanu, T.~Mikolov, and Y.~Bengio, ``On the difficulty of training
  recurrent neural networks,'' in \emph{International conference on machine
  learning}, 2013, pp. 1310--1318.

\bibitem{transformer}
A.~Vaswani, N.~Shazee, N.~Parmar, J.~Uszkorei, L.~Jone, A.~N. Gomez, Łukasz
  Kaiser, and I.~Polosukhin, ``Attention is all you need,'' \emph{arXiv}, no.
  1706.03762, Jun 2017.

\bibitem{roberta}
Y.~Liu, M.~Ott, N.~Goyal, J.~Du, M.~Joshi, D.~Chen, O.~Levy, M.~Lewis,
  L.~Zettlemoyer, and V.~Stoyanov, ``Roberta: A robustly optimized bert
  pretraining approach,'' \emph{arXiv}, no. 1907.11692, Jul 2019.

\bibitem{wordpiece}
M.~Schuster and K.~Nakajima, ``Japanese and korean voice search,'' in
  \emph{2012 IEEE International Conference on Acoustics, Speech and Signal
  Processing (ICASSP)}.\hskip 1em plus 0.5em minus 0.4em\relax IEEE, 2012, pp.
  5149--5152.

\bibitem{transformerXL}
Z.~Dai, Z.~Yang, Y.~Yang, W.~W. Cohen, J.~Carbonell, Q.~V. Le, and
  R.~Salakhutdinov, ``Transformer-xl: Attentive language models beyond a
  fixed-length context,'' \emph{arXiv}, no. 1901.02860, Jan 2019.

\bibitem{ulmfit}
Howard and Ruder, ``Universal language model fine-tuning for text
  classification,'' \emph{arXiv}, no. 1801.06146, 2018.

\bibitem{deepfeat}
Y.~B. Jason~Yosinski, Jeff~Clune and H.~Lipson, ``How transferable are features
  in deep neural networks?'' \emph{Advances in neural information processing
  systems}, pp. 3320--3328, 2014.

\bibitem{bertat}
O.~L. C. D.~M. Kevin~Clark, Urvashi~Khandelwal, ``What does bert look at? an
  analysis of bert’s attention,'' \emph{arXiv}, no. 1906.0434, Jun 2019.

\bibitem{bertclass}
Y.~X. X.~H. Chi~Sun, Xipeng~Qiu, ``How to fine-tune bert for text
  classification?'' \emph{arXiv}, no. 1905.05583, May 2019.

\end{thebibliography}

\end{document}